\documentclass[runningheads]{llncs}
\usepackage{graphicx}
\usepackage{floatflt}
\usepackage{amsmath,amssymb,amsfonts}
\usepackage{algorithm,algorithmic}
\usepackage{graphicx}
\usepackage{textcomp}
\usepackage{xcolor}
\usepackage{color}
\usepackage{colortbl}
\usepackage{tabularx,booktabs}
\usepackage{supertabular}
\usepackage{fancyhdr}
\usepackage{stfloats}
\usepackage{booktabs}
\usepackage{multicol}
\usepackage{multirow}
\usepackage{pgfplots}
\usepackage{pgfplotstable}
\usepackage{wrapfig}
\usepackage{sfmath}
\usepackage{todonotes}
\usepackage{makeidx}
\usepackage{soul}
\pgfplotsset{width=7cm,compat=1.8}
\usepackage{tikz}
\usetikzlibrary{shapes.geometric, arrows}
\usetikzlibrary{arrows.meta,
                chains,
                positioning}

\usepackage[nottoc]{tocbibind}
\usepackage[square,numbers]{natbib}
\begin{document}

\title{A Meta Survey of Quality Evaluation Criteria in Explanation Methods \thanks {This research is partly founded by the Swedish Knowledge Foundation through the Industrial Research School INSiDR.}\\ \ \\
}
\titlerunning{Quality Evaluation Criteria in Explanation Methods}
%

\author{Helena Löfström\inst{1,3}
\and Karl Hammar\inst{2} \and Ulf Johansson\inst{2}
}
\authorrunning{H. Löfström, et al.}
%
\institute{Dept. of Information Technology, University of Borås, Sweden\\ 
 \and Dept. of Computing, Jönköping University, Sweden\\
 \and Jönköping International Business School, Sweden\\
\email{\{helena.lofstrom, karl.hammar, ulf.johansson\}@ju.se}}

\maketitle              
\begin{abstract}
Explanation methods and their evaluation have become a significant issue in explainable artificial intelligence (XAI) due to the recent surge of opaque AI models in decision support systems (DSS). Since the most accurate AI models are opaque with low transparency and comprehensibility, explanations are essential for bias detection and control of uncertainty. There are a plethora of criteria to choose from when evaluating explanation method quality. However, since existing criteria focus on evaluating single explanation methods, it is not obvious how to compare the quality of different methods. This lack of consensus creates a critical shortage of rigour in the field, although little is written about comparative evaluations of explanation methods. In this paper, we have conducted a semi-systematic meta-survey over fifteen literature surveys covering the evaluation of explainability to identify existing criteria usable for comparative evaluations of explanation methods. The main contribution in the paper is the suggestion to use appropriate trust as a criterion to measure the outcome of the subjective evaluation criteria and consequently make comparative evaluations possible. We also present a model of explanation quality aspects. In the model, criteria with similar definitions are grouped and related to three identified aspects of quality; model, explanation, and user. We also notice four commonly accepted criteria (groups) in the literature, covering all aspects of explanation quality: Performance, appropriate trust, explanation satisfaction, and fidelity. We suggest the model be used as a chart for comparative evaluations to create more generalisable research in explanation quality.

\end{abstract}

\section{Introduction}
AI model-based \textit{Decision support systems} (DSS) have become increasingly popular due to their possibility of solving a variety of tasks, such as music recommendations or medical diagnosis. However, the highly accurate AI models lack both transparency and comprehensibility in their predictions, which has caused a new field to emerge; \textit{explainable artificial intelligence} (XAI). In this field, the goal is to explain the opaque AI models with the help of explanation methods. The research within the area is extensive, and a plethora of methods have been developed to meet the need \cite{arrieta20, Carvalho19, adadi2018peeking}. However, the interest in explainability has simultaneously led to confusion on numerous fronts; it is, e.g., not clear how to select or evaluate explanation methods \cite{murdoch19}. 

In efforts to solve the problems, several taxonomies have been created showing how and when to use different explanation methods. In parallel, many various criteria have also been proposed to evaluate the quality of the methods. However, the criteria focus on the evaluation of single explanation methods, and it is not evident how to use them for comparative evaluation \cite{murdoch19,chromik2020taxonomy, linardatos2021explainable}. The lack of formality and consensus on assessment of explanation quality create significant problems for comparative evaluation \cite{Carvalho19, zhang2018explainable}. 

Many studies in explanation methods focus on how users experience the explanations, which is an essential aspect of the explanation quality. Since the explanations are presented to humans, they are required to be understood by humans \cite{chromik2020taxonomy,adadi2018peeking}. However, it is not unusual to evaluate an explanation method only with a selected user group, with few participants, measuring the satisfaction with the explanations among those users \cite{chromik2020taxonomy, linardatos2021explainable}. When comparing the quality of explanation methods, the user experience is essential and how the user acts upon the explanations. The goal is to help the user detect when to trust the system's predictions and when not to trust them, i.e., if the user can gain an appropriate trust in the system with the help of explanations. There is, in other words, a need to be able to compare different methods based on more objective measurements than solely user experience.

However, there are no generally accepted evaluation criteria or threshold values to reach acceptable results for comparative evaluations. Instead, it is up to the individual researcher to decide which criterion to use and when it has reached a sufficient level. At the same time, there are many criteria to choose among, which makes it challenging to compare the results from different evaluations. Depending on the researcher's background, the criteria are also named differently, increasing the difficulties of comparison \cite{adadi2018peeking}. Comparative evaluations finally almost seem impossible when considering the focus on subjective user criteria, such as expectation and curiosity.

The contributions of this paper are: \begin{itemize}
\item We suggest measuring the outcome of the subjective evaluation criteria and thus make comparative evaluations possible.
\item We also present a model, identifying existing evaluation criteria from four different research areas, the aspect of quality they measure and how they relate to each other.
\end{itemize}

The remainder of this paper is structured as follows: the next section provides a brief summary of the concept of post hoc explanation methods. Section \ref{sec:method} outlines the literature study set-up, while the results and discussion are presented in Section \ref{sec:results}. The paper ends with some concluding remarks in Section \ref{sec:conclusions}.
 
\section{Post Hoc Explanations}
One of the reasons why evaluation of explanation methods is seen as complicated is the vagueness of the terms \textit{explainability} and \textit{interpretability}. Some authors take a serious stand against using the terms interchangeably, and some blend them. Several authors point to the no agreed-upon meaning and the consequences for the field when handling the term in a quasi-mathematical way  \cite{lipton2018, Carvalho19}. There are several attempts to create a unified definition of the concepts. In \cite{gunning19, arrieta20} \textit{explainability} is defined as a model's ability to make its functioning clearer to an audience. The definition is similar in \cite{das2020opportunities}, where it is seen as a way of verifying the output decision from an AI model. The terms are used separately in \cite{arrieta20}: 
\begin{itemize}
    \item Interpretability: A passive property of the model (the model \textit{is} interpretable), close to comprehensibility and transparency 
    \item Explainability: An active property (the explanation method \textit{explains to} the user). 
\end{itemize}

This study follows the same terminology as \cite{arrieta20}.

Research on explanations can, based on the definitions above, be divided into two main focus areas \cite{lundberg2017unified, moradi2021post}; transparency through inherently interpretable models (interpretability) and \textit{post-hoc} explanation methods for explaining opaque models (explainability).

\begin{figure}[H]
    \centering
    \vspace{-10pt}
    \includegraphics[width= 0.8\textwidth, height= 3 cm]{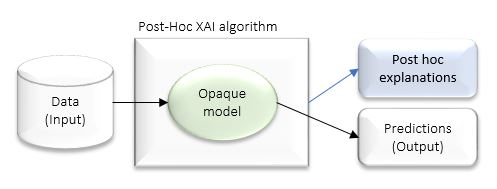}
    \caption{The extraction of Post-hoc explanations, inspired by \cite{arrieta20}}
    \label{fig:Helena_Lofstrom_ExM}
    \vspace{-10pt}
\end{figure}

Post-hoc explanation methods apply to the output of the underlying model and create a simplified and interpretable model based on the relation between feature values and the prediction (see figure \ref{fig:Helena_Lofstrom_ExM}). The relationships in the interpretable model are presented as explanations and can be, e.g., pixels in pictures, feature importance charts or words in texts, highlighting which features (pixels or words) that are important for the prediction. Explanations are intended to explain the model's strengths and weaknesses to the users, creating a possibility to identify erroneous predictions and an understanding of the model's rationale. Although the most typical situation with explanation methods is to explain the predictions of an opaque (so-called black box) model, explanation methods are also effective together with an inherently interpretable model, such as decision trees.

\section{Method} \label{sec:method}
The method in the study is a semi-systematic meta-survey. To ensure the quality of our research, we followed the methodology according to \cite{snyder2019literature} but also took into consideration \cite{webster2002analyzing}. The methodology from \cite{snyder2019literature} is divided into three primary steps: \textit{Designing} the review (including choice of search terms, databases, inclusion and exclusion criteria, which type of information to extract, and the type of analysis), \textit{conducting the review} with documentation of the process, and \textit{analysing} the results based on the choices made in the design. 
 \begin{wrapfigure}{R}{0.5\textwidth}
 \vspace{-10pt}
 \scalebox{0.5}
    \centering
    \begin{tikzpicture}
    [
        node distance = 3mm and 5mm,
        start chain = A going below,
        dot/.style = {circle, draw=white, very thick, fill=gray,minimum size=3mm},
        box/.style = {rectangle, text width=50mm,inner xsep=4mm, inner ysep=1mm,
        font=\sffamily\small\linespread{0.84}\selectfont,on chain},
    ]
    \begin{scope}[every node/.append style={box}
    ]
        \node 
        { 
                Total number of included surveys \\ 
        };
        \node 
        { 
                Phase 3: Citings \\
        };
        \node 
        { 
                Phase 2: References \\ 
        }; 
        \node 
        { 
                Second selection: Surveys\\ 
        };        
        \node 
        { 
               First Selection: Abstract reading \\ 
        };        
        \node 
        {
                Phase 1\\
                Explanation method (138)\\ 
                Trust (112)                \\      
        };        
    \end{scope}
    \draw[very thick, gray, {Triangle[length=4pt)]}-{Circle[length=3pt]},shorten <=-3mm, shorten >=-3mm]   
    (A-1.north west) -- (A-6.south west);
    \foreach \i [ count=\j] in {15,+3,+3,-27,-214, 250} 
    \node[dot,label=left:\i] at (A-\j.west) {};
\end{tikzpicture}
    \caption{The phases in the article selection (to be read bottom-up).}
    \label{fig:HelenaLofstrom_phase_method}  

\vspace{-28pt}

\end{wrapfigure}
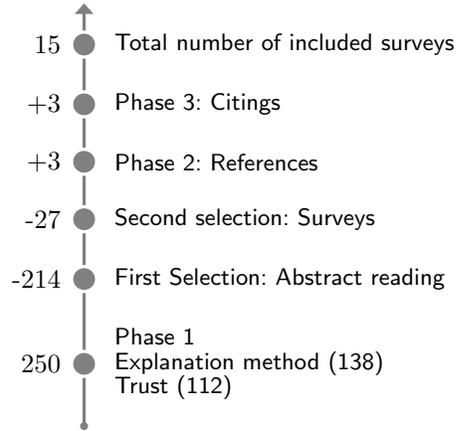

\subsection{Design}
The research in explanation methods is extensive, studied by various researchers within diverse disciplines. Due to this complexity, we used a semi-systematic literature meta-survey to cover several research areas and a larger number of articles. The literature was collected from the areas of \textit{Computer Science}, \textit{Social Science}, \textit{Business and Marketing Science} as well as \textit{Decision Science}. It is not unusual to use this type of multidisciplinary approach to provide an understanding of complex areas in semi-systematic literature methods \cite{snyder2019literature}. The method from \cite{snyder2019literature} was used with a content analysis of the included articles, looking for evaluation criteria, their definitions, usage, and quality threshold values. 

We used different databases with research articles (Scopus, Google Scholar, and Web of Science), resulting in similar articles. Several well-cited articles were also found in the ArXiv database, with over 1000 citations, where many were peer-reviewed papers. The articles from ArXiv are included due to their impact by the high amount of citations. From the initial literature review, two terms were most frequently used as keywords, and presumably not excluding any article:  \textit{Trust} and \textit{Explanation method}. We also used several other search terms in the initial literature research (evaluation criteria, evaluation, metrics, taxonomy). However, the criteria limited the number of results of articles and were excluded.
\begin{table}[H]
    \footnotesize
    \centering
    \begin{tabular} 
        {
                    >{\raggedright\arraybackslash}p{0.65\textwidth}
                    >{\raggedright\arraybackslash}p{0.10\textwidth}
                    >{\raggedright\arraybackslash}p{0.05\textwidth}
                    >{\raggedleft\arraybackslash}p{0.06\textwidth}
                    >{\raggedleft\arraybackslash}p{0.08\textwidth}
        }                    
        \toprule Title of Survey Article & Area & Ref. & Year &  NO arts.\\
       \toprule
        Towards a Rigorous Science of Interpretable Machine Learning &  IML & \cite{doshi2017towards} & 2017 & 51\\
        \addlinespace[3pt] 
        Explaining Explanations: An Overview of Interpretability of Machine Learning &  CS & \cite{gilpin2018explaining} & 2018 & 87\\
        \addlinespace[3pt]  
        Peeking inside the Black box & CS & \cite{adadi2018peeking} & 2018 & 180 \\
        \addlinespace[3pt]      
        Explainable Recommendations: A Survey and New Perspectives & RSS & \cite{zhang2018explainable} & 2018 & 178\\
        \addlinespace[3pt]
        A multidisciplinary survey and framework for design and \newline evaluation of explainable AI systems & XAI & \cite{mohseni2018multidisciplinary} & 2018 & 226\\
        \addlinespace[3pt]      
        Metrics for Explainable AI: Challenges and Prospects & HCI & \cite{hoffman2018metrics} & 2019 & 161\\
        \addlinespace[3pt]
        DARPA's Explainable Artificial Intelligence Program & XAI & \cite{gunning19} & 2019 & 30\\
        \addlinespace[3pt]
        Designing Theory-Driven User-Centric Explainable AI & Human factors & \cite{wang2019Designing} & 2019 & 109\\
        \addlinespace[3pt]        
        Machine learning Interpretability: \newline A survey on methods and metrics & CS,\newline HCI & \cite{Carvalho19} & 2019 & 153\\
        \addlinespace[3pt]
        Explanation in Human AI systems: A Literature Meta-Review Synopsis of Key Ideas and Publications and Bibliography for Explainable AI & XAI & \cite{mueller19} & 2019 & 739\\
        \addlinespace[3pt]
        A Taxonomy for Human Subject Evaluating of Black Box Explanations in XAI & HCI & \cite{chromik2020taxonomy} & 2020 & 33\\
        \addlinespace[3pt]     
        Opportunities and Challenges in Explainable Artificial Intelligence (XAI) & XAI & \cite{das2020opportunities} & 2020 & 119\\
        \addlinespace[3pt]        
        Explainable Artificial Intelligence (XAI): Concepts, Taxonomies, Opportunities and challenges toward responsible AI  & HCI &  \cite{arrieta20} & 2021 & 426\\
        \addlinespace[3pt]
        Explainable AI: A review of Machine Learning Interpretability Methods & XAI, IML & \cite{linardatos2021explainable} & 2021 & 165\\
        \addlinespace[3pt]
        Evaluating the Quality of Machine Learning Explanations: A survey on Methods and Metrics & IML & \cite{zhou2021evaluating}& 2021 & 93\\
        \midrule
        \multicolumn{4}{l}{Theoretical minimum number of included articles (maximum overlap): } & 739\\
        \multicolumn{4}{l}{Theoretical maximum number of included articles (minimum overlap): } & 2, 750\\
        \bottomrule\\
       \end{tabular}
        \caption{The included surveys.}
        \label{HelenaLofstrom_tab_Included_A}
         \hfill
  \normalsize
  \end{table}
\subsection{Conduct}
When conducting the meta-survey, three phases were considered (see also figure \ref{fig:HelenaLofstrom_phase_method}):
\begin{enumerate}
    \item A main survey, with the primary keywords of either \textit{explanation method} or \textit{trust}.
    \begin{itemize}
        \item Two hundred fifty (250) articles were found. One hundred thirty-eight (138) articles with the keyword explanation method, and one hundred twelve (112) articles with the keyword trust. In this stage, the search was not limited to surveys since the keywords do not always indicate if it is a survey.
    \item The abstracts were read, and those with a focus that coincided with the evaluation of explanation methods were chosen. \newline
    Two hundred fourteen (214) articles were excluded since they did not study evaluation of explanation methods or explanation quality criteria, resulting in 36 articles.
    \item The abstracts were reread, and among the 36 articles, only surveys were chosen (taxonomies are in this paper regarded as a type of survey), which resulted in the exclusion of 27 additional articles.
    \end{itemize}
    Phase 1 resulted in a total of 9 (nine) surveys from four different research areas.
    \item The list of references in the resulting nine surveys were studied, and three additional surveys were found. Phase 2 resulted in a total of 12 surveys.
    \item Articles that cited the surveys were also studied. Since at least one of the surveys were cited over 1000 times, a limited reverse snowballing method was used. Three (3) more surveys were found, and phase 3 resulted in 14 unique surveys.
\end{enumerate}

It was presumed that since several of the surveys covered the same years (see table \ref{HelenaLofstrom_tab_Included_A}), there would to some extent be overlapping articles. With a minimum (i.e., zero) overlap, this could theoretically result in a maximum of 2 750 articles and a minimum of 739 articles (maximum overlap) (see table \ref{HelenaLofstrom_tab_Included_A}). The actual number of articles is probably somewhere between the minimum and maximum. Several of the included surveys singled out similar articles as significant for the area of research, \cite{lipton2018, hoffman2018metrics, doshi2017towards, ribeiro2016should, gilpin2018explaining}. These were included in the meta-survey as articles of particular interest.

\subsection{Analysis}
 When the final sample was selected, the surveys were analysed to find evaluation criteria (definitions, usage, and quality threshold value). Each evaluation criterion was initially documented separately and then grouped based on definition (see table \ref{HelenaLofstrom_tab_Explanation_characteristics}).
 The analysis also included several additional steps:
\begin{itemize}
    \item The surveys were studied to find the different aspects of explanation (method) quality. The analysis of explanation aspects focused on finding i) how the 
    \begin{table}[H]
    \centering
    \footnotesize
    \begin{tabular} 
        {
                    >{\raggedright\arraybackslash}p{0.30\textwidth}
                    >{\raggedright\arraybackslash}p{0.4\textwidth}
                    >{\centering\arraybackslash}p{0.15\textwidth}
                    >{\raggedleft\arraybackslash}p{0.15\textwidth}
                    }
       \toprule Criterion & Description & Explanation \newline aspect & Survey\\
       \toprule
        \textbf{Performance},\newline Accuracy, Accountability, Confidentiality, Certainty, Confidence & The performance of the model. & Model & \cite{arrieta20} \cite{chromik2020taxonomy} \cite{adadi2018peeking} \cite{hoffman2018metrics} \cite{wang2019Designing} \cite{mohseni2018multidisciplinary} \cite{gunning19}\cite{murdoch19} \cite{Carvalho19}  \cite{das2020opportunities} \cite{linardatos2021explainable} \\
       \addlinespace[7pt]
       \textbf{Fairness} \newline Bias (detection) & The degree to which predictions are unbiased and do not implicitly or explicitly discriminate against protected groups. & Model & \cite{arrieta20} \cite{doshi2017towards} \cite{chromik2020taxonomy} \cite{das2020opportunities} \cite{adadi2018peeking}\cite{mohseni2018multidisciplinary} \cite{wang2019Designing} \cite{gunning19} \cite{zhang2018explainable}\\ 
       \addlinespace[7pt]
       \textbf{Privacy} & Sensitive information in the model is protected. & Model & \cite{Carvalho19},\cite{arrieta20} \cite{doshi2017towards} \cite{das2020opportunities}\\
       \addlinespace[7pt]
       \textbf{Reliability}, \newline Robustness & How much small changes in the input causes changes in the prediction. & Model &\cite{doshi2017towards}\cite{arrieta20}\cite{chromik2020taxonomy} \cite{hoffman2018metrics}\\
       \addlinespace[7pt]
        \textbf{Identity},\newline Stability & Identical instances should have identical explanations. & Expl. & \cite{Carvalho19}\\
       \addlinespace[7pt]
       \textbf{Separability}, Consistency & Non-identical instances should not have identical explanations. & Expl. & \cite{Carvalho19}\\
       \addlinespace[7pt]
       \textbf{Novelty} & The instance should not come from a region in instance space that is far from the training data. & Expl. & \cite{Carvalho19}\\
       \addlinespace[7pt]
       \textbf{Representativeness} & How many instances that are covered by the explanations. Explanations can cover the entire model. & Expl. & \cite{Carvalho19}\\
       \addlinespace[7pt]
       \textbf{Fidelity},\newline Description accuracy, \newline  Feature importance/ relevance, BAM\footnote{Benchmarketing Attribution Methods}, Faithfulness, Monotonicity  & Measures the accuracy of the explanation against the model. & Expl. & \cite{hoffman2018metrics} \cite{Carvalho19} \cite{arrieta20} \cite{das2020opportunities} \cite{adadi2018peeking} \cite{wang2019Designing}\cite{mueller19}\\
       \addlinespace[7pt]
       \textbf{Appropriate Trust}\newline Calibrated Trust, Trust, Reliance & How much a user can distinguish between correct and erroneous predictions and act based on it. & User & \cite{Carvalho19}\cite{arrieta20}\cite{chromik2020taxonomy}\cite{hoff2015trust} \cite{doshi2017towards}\cite{das2020opportunities}\cite{adadi2018peeking}\cite{hoffman2018metrics}\newline\cite{wang2019Designing}\cite{mohseni2018multidisciplinary}\cite{gunning19}\cite{zhang2018explainable}\\
       \addlinespace[7pt]
       \textbf{Explanation Satisfaction},\newline Satisfaction Scale, \newline Comprehensibility,\newline Explanation Goodness,\footnote{In \cite{hoffman2018metrics} there is two different scales to measure the Explanation Satisfaction Scale and the Explanation Goodness Checklist, however they are in this paper seen as the same criterion} Causability,\newline System Causability Scale & The degree of how much the users feel they understand the system, the explanations, and the user interface. & User & \cite{hoffman2018metrics} \cite{arrieta20} \cite{chromik2020taxonomy}  \cite{mueller19} \cite{mohseni2018multidisciplinary} \cite{gunning19} \cite{zhang2018explainable}\cite{das2020opportunities}\cite{doshi2017towards} \cite{hoff2015trust}\\
       \bottomrule\\
       \end{tabular}
       \caption{The identified explanation criteria.}
       \label{HelenaLofstrom_tab_Explanation_characteristics}
       \hfill\\
    \end{table}
    \normalsize    
\clearpage
    explanation criteria were collected, ii) if the criteria were \textit{objective} (machine outcome) or \textit{subjective} (human outcome), and iii) what the criteria were supposed to measure. 
    \item Based on i, ii, and iii, the criteria were grouped on the aspect of explanation quality.
    \item One of the meta survey's goals was to find generally accepted evaluation criteria for comparative evaluations of explanation methods. To identify the most commonly accepted criteria, we compared in how many surveys each criterion was mentioned as necessary for evaluations. Of course, the number of times a criterion was mentioned could be affected because the criteria were grouped. However, it indicated the usage of the criteria. 
\end{itemize}

\begin{figure}[H]
    \vspace{-15pt}
    \pgfplotsset{width=7cm, height = 5.5cm, compat=1.15}
    \begin{center}

        \begin{tikzpicture}[scale=0.9]
        \begin{axis}
            [
            xbar,
            y axis line style = { opacity = 0 },
            axis x line       = none,
            tickwidth         = 0pt,
            ytick=data,
            enlarge y limits=0.05,
            enlarge x limits  = 0.02,
            width=0.7\textwidth,
            bar width=2mm,
            xlabel={\#Articles},
            symbolic y coords={Performance,Fairness,Privacy, Reliability, Identity, Separability, Novelty, Representativeness,Fidelity, Appropriate Trust,Explanation Satisfaction},
            nodes near coords, 
            ]
            \addplot coordinates{(11,Performance)(9,Fairness)(4,Privacy)(4,Reliability)(1,Identity)(1,Separability)(1,Novelty)(1,Representativeness)(7,Fidelity)(12,Appropriate Trust)(10,Explanation Satisfaction)};
            \smallskip
            \end{axis}
            \end{tikzpicture}
        \caption{Number of surveys mentioning the evaluation criteria (groups) as important.}
        \label{fig:Helena_Lofstrom_fig_comp_cr}
    \end{center}          
\end{figure}
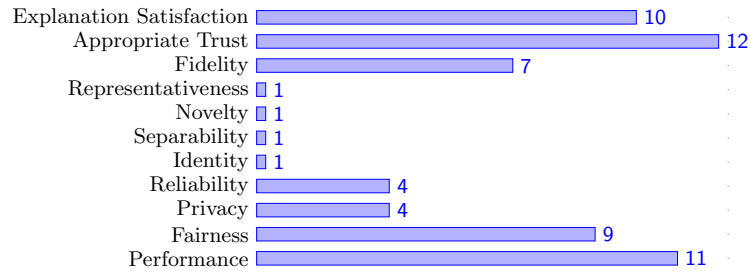

\section{Results and Discussion} \label{sec:results}
\begin{figure}[H]
    \centering
    \includegraphics[width = \textwidth, height= 9 cm]{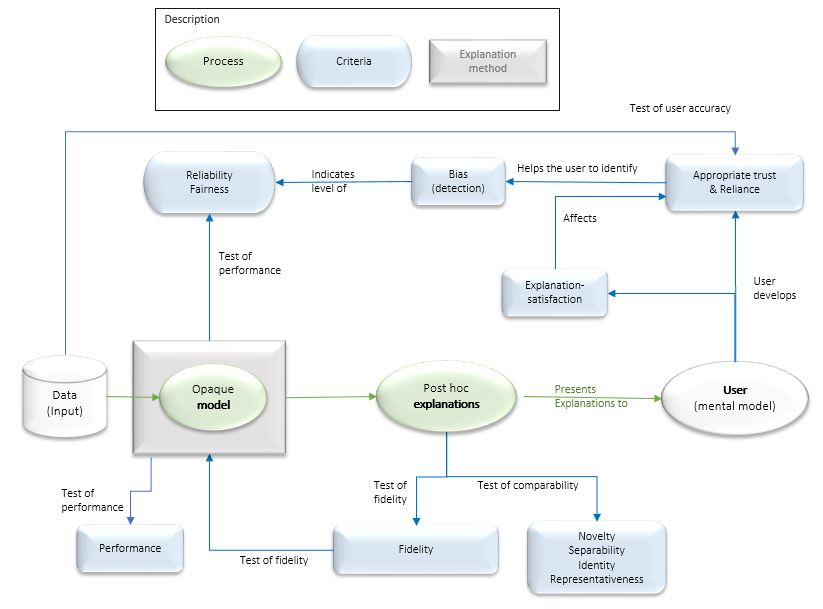}
    \caption{High-level model of the criteria groups in the different aspects of explanation quality.}
     \vspace{-10pt}
    \label{fig:Helena_Lofstrom_General_Overview}
\end{figure}

It is generally accepted that the evaluation of explanations is of a multidimensional character, having primarily two aspects of explanation quality: \textit{user} and \textit{explanation} \cite{doshi2017towards,murdoch19,gilpin2018explaining}. Since the explanation quality is intimately connected to the underlying model and its level of accuracy, it is also essential to consider the quality of \textit{the model} \cite{hoffman2018metrics,Carvalho19}. In that sense, the quality of explanations is dependent on all three aspects; model, explanation, and user (see figure \ref{fig:Helena_Lofstrom_General_Overview}).  
All three aspects are necessary to consider but likewise essential to distinguish. The type and goal of the evaluation will change depending on the type of explanation. An evaluation trying, e.g., to catch the change in the user's mental model, will differ from the evaluation trying to catch the model's level of stability.
The explanations should be presented and understood by humans, but at the same time, it is not possible to evaluate explanation methods based on an entirely subjective outcome \cite{doshi2017towards,mueller19}. The division between the user and the explanation aspect is also referred in the literature as \textit{measurements} and \textit{metrics} \cite{arrieta20, doshi2017towards}.

In line with earlier research, we found that many evaluation criteria had various notions (names) in different studies. Based on the findings in the literature,

we grouped the criteria on how they were defined: how the criteria were \textit{collected}, if the criteria were  \textit{objective} or \textit{subjective}, and what the criteria were supposed to \textit{measure}. The most common or well-established names in the surveys were used as the name of the criteria (groups). Some of the criteria were vaguely described, lacking in usage description. Although several other criteria could be used in an evaluation and were connected to the resulting criteria, they were excluded due to the lack of (method) description of how to use or quantify them in evaluations. After the analysis, we identified eleven criteria (see table \ref{HelenaLofstrom_tab_Explanation_characteristics}) possible to use with some form of measured value. 
No threshold values to reach an acceptable quality level could be found, which cause another challenge for comparative evaluations. Without a threshold value for an acceptable level of quality, it is possible to compare if an explanation method is, e.g., creating a higher level of appropriate trust than another method. However, the comparison does not indicate if any of the two methods have reached an acceptable level of appropriate trust.

\subsection{The User Aspect}
The majority of the criteria in the user aspect are subjective and challenging to measure. However, since changes in the subjective criteria cause the user's \textit{mental model} to change during the usage of the system, we consider the mental model as a container of these criteria and not a criterion of its own \cite{mueller19,arrieta20,hoffman2018metrics, mohseni2018multidisciplinary,gunning19}. The resulting four criteria connected to the user aspect were:\\
\textbf{Trust}: a highly used criterion and one of the absolute most important criteria for success. It is subjective and therefore challenging to measure \cite{doshi2017towards, gunning19, Carvalho19, hoff2015trust,zhou2021evaluating}. However, one fundamental question is whether the users distinguish erroneous predictions and how they react. Some form of metrics is required that also measures the outcome of the system, the human-machine effectiveness.
\begin{itemize}
    \item \textit{Trust} a highly used criterion, and several of the surveys mention it as one of the absolute most important criteria for success. However, many surveys also highlight that it is subjective or user-centric, and therefore challenging to measure \cite{doshi2017towards, gunning19, Carvalho19, hoff2015trust,zhou2021evaluating}. The literal meaning of trust is the degree to which the user agrees with the model. If a user trusts a model entirely, the user accepts all incorrect and correct predictions and has 100\% trust in the model. Total trust in a model is only adequate if the model is 100 \% accurate. Otherwise, it is called \textit{misuse} and is hazardous if important decisions are to be taken based on the predictions. The users should not trust the predictions of the model more than its accuracy. A system where the user's trust is higher than the model's accuracy could, in other words, be a very inefficient DSS. Even though the users find the explanations highly satisfying, straightforward, and if the explanations create curiosity to find more knowledge, one fundamental question is whether the users distinguish erroneous predictions and react. Some form of metrics is required that also measures the outcome of the system, the human-machine effectiveness. 
    \item \textit{Appropriate trust}: Trust may also be defined as an attitude that is formed by information about the model and previous experiences. The attitude creates intentions whether to rely on the model or not, resulting in a behaviour, or \textit{reliance} \cite{Ekman17, gunning19}. Under this definition trust is seen in the literature as directly connected to the quality of post hoc explanations, in the form of \textit{appropriate trust} \cite{mcdermott2019practical}. An explanation method could, in other words, create a high trust but a lower appropriate trust. A good mental model is seen as a requirement for developing appropriate trust in the model \cite{hoff2015trust}, i.e. as could be seen in the model (see figure \ref{fig:Helena_Lofstrom_General_Overview} the criterion is one of the outcomes of the user's mental model, consisting of the user aspect (human trust in the system) and the explanation aspect (the output of the system) \cite{mcdermott2019practical, doshi2017towards}. Appropriate trust is within the XAI an accepted evaluation criterion to measure the quality of explanations \cite{Carvalho19,arrieta20,chromik2020taxonomy,hoff2015trust,doshi2017towards,das2020opportunities,adadi2018peeking,hoffman2018metrics,wang2019Designing,mohseni2018multidisciplinary,gunning19,zhang2018explainable}. When identifying the level of appropriate trust, the users try to identify the correct and erroneous predictions in a binary choice situation. The ideal situation is a 1:1 situation where all the correct predictions and all the erroneous predictions are identified. However, in none of the surveys could we find any threshold value for this criterion. The criterion could be defined as the users ability to \textit{“[not] follow an [in]correct recommendation”} \cite{gunning19, yang20}. All other cases lead to either \textit{misuse} (overuse) or \textit{disuse} (under-use). Appropriate trust could, in other words, be defined as the user's accuracy; to what degree the users act according to the data.

    \item \textit{Explanation satisfaction}: to what extent an explainable user interface or an explanation is suitable for the intended purpose \cite{holzinger2020measuring}. This criterion is the outcome of the mental model, together with appropriate trust (see figure \ref{fig:Helena_Lofstrom_General_Overview}). The evaluation is conducted with Likert scales, with questions similar to, e.g., if the user understands the explanations and finds them relevant. 
    \item \textit{Bias detection} is often seen as one of the desired outcome of explanations, when the user is able to distinguish erroneous predictions \cite{wang2019Designing,gunning19,zhang2018explainable,das2020opportunities}. However, \cite{das2020opportunities} defines the term as more of a data set desiderata. The criterion could be seen as the joint outcome of the user's mental model, together with appropriate trust, creating reliance (see figure \ref{fig:Helena_Lofstrom_General_Overview}). When able to detect the erroneous predictions and in that way also identifying the frequency of errors, the user could be able to identify the general patterns of bias in the system.
    \item The user aspect of explanation quality also includes four criteria that we found no formulated method description and described as very challenging to measure: \textit{expectation}, \textit{curiosity}, \textit{cognition} and \textit{context knowledge} \cite{hoffman2018metrics}.
\end{itemize}

\subsection{The Explanation Aspect}
The \textit{Explanation} aspect of quality is placed between the model aspect and the user aspect, focusing on the evaluations of the explanations. It is sometimes referred to as the \textit{user-machine performance}, highlighting the relation between the machine and the human. The criteria are objective but can include humans in the evaluation. Five criteria were identified in this aspect:
\begin{itemize}
    \item \textbf{Fidelity}: reflects how accurately the explanation method mirrors the underlying model \cite{moradi2021post, hoffman2018metrics, Carvalho19, arrieta20, das2020opportunities, adadi2018peeking, wang2019Designing, mueller19}. The fidelity could also be used to measure the difference between description accuracy given by the system and the description accuracy assessed by the user, including a more subjective perspective \cite{das2020opportunities}.
     \item \textbf{Identity, Separability, Novelty, and Representativeness}: in \cite{Carvalho19} the authors suggests several new metrics for evaluation of explanations. The criteria are related to fidelity and catch the explanations' correctness, although they can be calculated without the inclusion of humans. The three first criteria compare the explanations between different instances; i) identical instances should have identical explanations, ii) non-identical instances should not have identical explanations, and iii) the instance should not come from a region in instance space far from the training data. Finally, representativeness measures how many instances are covered by the explanation. 
\end{itemize}

\subsection{The Model aspect}
In the model aspect of explanation quality, the output from the user aspect, or mental model, are fed back to the system via the criteria, \textit{appropriate trust} and \textit{reliance} (see figure \ref{fig:Helena_Lofstrom_General_Overview}). In \cite{hoffman2018metrics, zhou2021evaluating} the authors point out how the system's performance, the evaluation of the user's performance, and the performance of the human-machine system cannot be neatly divorced from each other. Several evaluation criteria are connected to the underlying model's aspect of quality:
\begin{itemize}
    \item \textit{Performance} is a group of criterion that, together with appropriate trust, is most often referred to in the surveys (see figure \ref{fig:Helena_Lofstrom_fig_comp_cr}) as important to the quality of the explanations and denotes the level of correctness of the model \cite{arrieta20,chromik2020taxonomy,adadi2018peeking,hoffman2018metrics,wang2019Designing,mohseni2018multidisciplinary,Carvalho19,gunning19,murdoch19,das2020opportunities,linardatos2021explainable, zhou2021evaluating}. 
    In \cite{zhou2021evaluating} several different evaluation criteria for the model is presented connected to the performance of the model, although the authors highlight the criterion \textit{soundness of fidelity}.
    \item \textit{Fairness}, also called unbiasedness is the opposite to bias. However, they measure the same aspect of quality: to what degree the model have general patterns of errors. A model without bias is unbiased \cite{doshi2017towards}, i.e., have a high level of fairness. The level of bias indicates the level of fairness and also the level of reliability of the model (see figure \ref{fig:Helena_Lofstrom_General_Overview})
    \item \textit{Reliability} is close to the criterion Accuracy, since it signals a confidence measure of the model to the user in a specific situation  \cite{murdoch19,Carvalho19,arrieta20,das2020opportunities,chromik2020taxonomy,hoffman2018metrics,doshi2017towards}. However, it could be argued that these two criteria in reality are synonyms to each other, specifically when they are referred to the very similar names of \textit{certainty} and \textit{confidence}. The latter are simultaneously not seldom also defined as the \textit{stability} of the model, indicating that it could be evaluated from a different angle. This vagueness is not unusual to find, which is one reason why the area of research needs commonly accepted definitions of the terms. In \cite{Carvalho19} the authors highlight an interesting aspect of the certainty criterion; that many ML models only provide prediction values which do not include statements about the model's confidence on the correctness of the prediction.
    \item \textit{Privacy} This criterion has no well defined method and is referred to that the model protects sensitive information in the data \cite{doshi2017towards}.
\end{itemize}

\subsection{Discussion}
Comparative evaluations of explanation methods are challenging, and a general computational benchmark across all possible explanation methods is seen in \cite{zhou2021evaluating} as unlikely to be possible due to the subjective characteristics. However, it is generally accepted that the mental model affects the level of appropriate trust and reliance in the system. If considering the mental model as a container of the subjective criteria, the outcome of the mental model would be possible to measure through the criteria appropriate trust (see, e.g., \cite{hoffman2018metrics}). Although appropriate trust does not explicitly answer how the user experiences the explanations, it demonstrates if they fulfil one of the most crucial goals of explanation methods; if the user can detect correct and erroneous predictions. By measuring the outcome of the mental model through appropriate trust, we get an objective metric for the quality of the user aspect and create possibilities for comparative evaluations of explanation methods.

It is essential to highlight that we do not consider the subjective criteria unnecessary or unimportant to measure. In contrast, we acknowledge them as crucial for quality. When evaluating a single explanation method, it could be vital to follow subjective criteria changes. However, if the evaluation intends to gain comparable results, we recommend using the appropriate trust criteria.

\section{Conclusion} \label{sec:conclusions}
This paper conducted a semi-systematic meta-survey over fifteen surveys to find commonly accepted evaluation criteria, with a well-defined method possible to use in comparative evaluations. The field was found to focus on human-in-the-loop evaluations, creating severe challenges for quality comparisons.

The major contribution in the paper is the suggestion of using the criterion \textit{appropriate trust} as an outcome metric of the subjective criteria in the mental model, overcoming the problems with comparative evaluations. We also present a high-level model of explanation method quality, identifying three aspects of quality: \textit{model}, \textit{explanation}, and \textit{user}. The quality of an explanation method is suggested to be a composition of all three aspects. We also identified eleven evaluation criteria groups in different aspects of quality and their relation in the meta-survey. Four of the criteria were mentioned in more than half of the surveys as necessary for the quality of an explanation method: \textit{performance}, \textit{appropriate trust}, \textit{explanation satisfaction}, and \textit{fidelity}. These criteria cover all three aspects of explanation quality, and we suggest that they are used when researchers want to make their evaluations comparable.

\bibliography{Helena_lofstrom_main.bib}

\begin{thebibliography}{10}

\bibitem{arrieta20}
Alejandro~Barredo Arrieta, Natalia D{\'\i}az-Rodr{\'\i}guez, Javier Del~Ser,
  Adrien Bennetot, Siham Tabik, Alberto Barbado, Salvador Garc{\'\i}a, Sergio
  Gil-L{\'o}pez, Daniel Molina, Richard Benjamins, et~al.
\newblock Explainable artificial intelligence (xai): Concepts, taxonomies,
  opportunities and challenges toward responsible ai.
\newblock {\em Information Fusion}, 58:82--115, 2020.

\bibitem{Carvalho19}
Diogo~V. Carvalho, Eduardo~M. Pereira, and Jaime~S. Cardoso.
\newblock Machine learning interpretability: A survey on methods and metrics.
\newblock {\em Electronics}, 8:832, 2019.

\bibitem{adadi2018peeking}
Amina Adadi and Mohammed Berrada.
\newblock Peeking inside the black-box: A survey on explainable artificial
  intelligence (xai).
\newblock {\em IEEE Access}, 6:52138--52160, 2018.

\bibitem{murdoch19}
W~James Murdoch, Chandan Singh, Karl Kumbier, Reza Abbasi-Asl, and Bin Yu.
\newblock Definitions, methods, and applications in interpretable machine
  learning.
\newblock {\em Proceedings of the National Academy of Sciences},
  116(44):22071--22080, 2019.

\bibitem{chromik2020taxonomy}
Michael Chromik and Martin Schuessler.
\newblock A taxonomy for human subject evaluation of black-box explanations in
  xai.
\newblock In {\em ExSS-ATEC@ IUI}, 2020.

\bibitem{linardatos2021explainable}
Pantelis Linardatos, Vasilis Papastefanopoulos, and Sotiris Kotsiantis.
\newblock Explainable ai: A review of machine learning interpretability
  methods.
\newblock {\em Entropy}, 23(1):18, 2021.

\bibitem{zhang2018explainable}
Yongfeng Zhang and Xu~Chen.
\newblock Explainable recommendation: A survey and new perspectives.
\newblock {\em arXiv preprint arXiv:1804.11192}, 2018.

\bibitem{lipton2018}
Zachary~C Lipton.
\newblock The mythos of model interpretability.
\newblock {\em Queue}, 16(3):31--57, 2018.

\bibitem{gunning19}
David Gunning and David~W Aha.
\newblock Darpa’s explainable artificial intelligence program.
\newblock {\em AI Magazine}, 40(2):44--58, 2019.

\bibitem{das2020opportunities}
Arun Das and Paul Rad.
\newblock Opportunities and challenges in explainable artificial intelligence
  (xai): A survey.
\newblock {\em arXiv preprint arXiv:2006.11371}, 2020.

\bibitem{lundberg2017unified}
Scott~M Lundberg and Su-In Lee.
\newblock A unified approach to interpreting model predictions.
\newblock In {\em Proceedings of the 31st international conference on neural
  information processing systems}, pages 4768--4777, 2017.

\bibitem{moradi2021post}
Milad Moradi and Matthias Samwald.
\newblock Post-hoc explanation of black-box classifiers using confident
  itemsets.
\newblock {\em Expert Systems with Applications}, 165:113941, 2021.

\bibitem{snyder2019literature}
Hannah Snyder.
\newblock Literature review as a research methodology: An overview and
  guidelines.
\newblock {\em Journal of business research}, 104:333--339, 2019.

\bibitem{webster2002analyzing}
Jane Webster and Richard~T Watson.
\newblock Analyzing the past to prepare for the future: Writing a literature
  review.
\newblock {\em MIS quarterly}, pages xiii--xxiii, 2002.

\bibitem{doshi2017towards}
Finale Doshi-Velez and Been Kim.
\newblock Towards a rigorous science of interpretable machine learning.
\newblock {\em arXiv preprint arXiv:1702.08608}, 2017.

\bibitem{gilpin2018explaining}
Leilani~H Gilpin, David Bau, Ben~Z Yuan, Ayesha Bajwa, Michael Specter, and
  Lalana Kagal.
\newblock Explaining explanations: An overview of interpretability of machine
  learning.
\newblock In {\em 2018 IEEE 5th International Conference on data science and
  advanced analytics (DSAA)}, pages 80--89. IEEE, 2018.

\bibitem{mohseni2018multidisciplinary}
Sina Mohseni, Niloofar Zarei, and Eric~D Ragan.
\newblock A multidisciplinary survey and framework for design and evaluation of
  explainable ai systems.
\newblock {\em arXiv}, pages arXiv--1811, 2018.

\bibitem{hoffman2018metrics}
Robert~R Hoffman, Shane~T Mueller, Gary Klein, and Jordan Litman.
\newblock Metrics for explainable ai: Challenges and prospects.
\newblock {\em arXiv preprint arXiv:1812.04608}, 2018.

\bibitem{wang2019Designing}
Danding Wang, Qian Yang, Ashraf Abdul, and Brian~Y. Lim.
\newblock Designing theory-driven user-centric explainable ai.
\newblock In {\em Proceedings of the 2019 CHI Conference on Human Factors in
  Computing Systems}, CHI '19, page 1–15, New York, NY, USA, 2019.
  Association for Computing Machinery.

\bibitem{mueller19}
Shane~T Mueller, Robert~R Hoffman, William Clancey, Abigail Emrey, and Gary
  Klein.
\newblock Explanation in human-ai systems: A literature meta-review, synopsis
  of key ideas and publications, and bibliography for explainable ai.
\newblock {\em arXiv preprint arXiv:1902.01876}, 2019.

\bibitem{zhou2021evaluating}
Jianlong Zhou, Amir~H Gandomi, Fang Chen, and Andreas Holzinger.
\newblock Evaluating the quality of machine learning explanations: A survey on
  methods and metrics.
\newblock {\em Electronics}, 10(5):593, 2021.

\bibitem{ribeiro2016should}
Marco~Tulio Ribeiro, Sameer Singh, and Carlos Guestrin.
\newblock Why should i trust you?" explaining the predictions of any
  classifier.
\newblock In {\em Proceedings of the 22nd ACM SIGKDD international conference
  on knowledge discovery and data mining}, pages 1135--1144, 2016.

\bibitem{hoff2015trust}
Kevin~Anthony Hoff and Masooda Bashir.
\newblock Trust in automation: Integrating empirical evidence on factors that
  influence trust.
\newblock {\em Human factors}, 57(3):407--434, 2015.

\bibitem{Ekman17}
Fredrick Ekman, Mikael Johansson, and Jana Sochor.
\newblock Creating appropriate trust in automated vehicle systems: A framework
  for hmi design.
\newblock {\em IEEE Transactions on Human-Machine Systems}, 48(1):95--101,
  2017.

\bibitem{mcdermott2019practical}
Patricia~L McDermott and Ronna N~ten Brink.
\newblock Practical guidance for evaluating calibrated trust.
\newblock In {\em Proceedings of the Human Factors and Ergonomics Society
  Annual Meeting}, volume~63, pages 362--366. SAGE Publications Sage CA: Los
  Angeles, CA, 2019.

\bibitem{yang20}
Fumeng Yang, Zhuanyi Huang, Jean Scholtz, and Dustin~L Arendt.
\newblock How do visual explanations foster end users' appropriate trust in
  machine learning?
\newblock In {\em Proceedings of the 25th International Conference on
  Intelligent User Interfaces}, pages 189--201, 2020.

\bibitem{holzinger2020measuring}
Andreas Holzinger, Andr{\'e} Carrington, and Heimo M{\"u}ller.
\newblock Measuring the quality of explanations: the system causability scale
  (scs).
\newblock {\em KI-K{\"u}nstliche Intelligenz}, pages 1--6, 2020.

\end{thebibliography}

\end{document}